\documentclass[sigconf]{acmart}

\usepackage{multirow}
\usepackage{graphicx}
\newenvironment{mybox}
    {
    \begin{tabular}{p{0.43\textwidth}}
    \\
    }
    { 
    \\
    \end{tabular} 
    }

\AtBeginDocument{%
  \providecommand\BibTeX{{%
    \normalfont B\kern-0.5em{\scshape i\kern-0.25em b}\kern-0.8em\TeX}}}

\setcopyright{acmcopyright}
\copyrightyear{2023} 
\acmYear{2023} 
\setcopyright{acmlicensed}\acmConference[ICAIL 2023]{Nineteenth International Conference on Artificial Intelligence and Law}{June 19--23, 2023}{Braga, Portugal}
\acmBooktitle{Nineteenth International Conference on Artificial Intelligence and Law (ICAIL 2023), June 19--23, 2023, Braga, Portugal}
\acmPrice{15.00}
\acmDOI{10.1145/3594536.3595170}
\acmISBN{979-8-4007-0197-9/23/06}

\begin{document}

\title{Legal Syllogism Prompting: Teaching Large Language Models for Legal Judgment Prediction}

\author{Cong Jiang}
\affiliation{%
  \institution{Peking University Law School}
  \institution{Institute for Artificial Intelligence, Peking University}
  \city{Beijing}
  \country{China}
  \\
  \institution{PKU-WUHAN Institute for Artificial Intelligence}
  \city{Wuhan}
  \country{Hubei}
  \country{China}}
\email{jiangcong@pku.edu.cn}

\author{Xiaolei Yang}
\affiliation{%
  \institution{PKU-WUHAN Institute for Artificial Intelligence}
  \city{Wuhan}
  \country{Hubei}
  \country{China}
  \\
  \institution{Peking University Law School}
  \institution{Institute for Artificial Intelligence, Peking University}
  \city{Beijing}
  \country{China}
}
\email{yangxiaolei@pku.edu.cn}

\renewcommand{\shortauthors}{Jiang and Yang}

\begin{abstract}
Legal syllogism is a form of deductive reasoning commonly used by legal professionals to analyze cases.  In this paper, we propose legal syllogism prompting (LoT), a simple prompting method to teach large language models (LLMs) for legal judgment prediction. LoT teaches only that in the legal syllogism the major premise is law, the minor premise is the fact, and the conclusion is judgment. Then the models can produce a syllogism reasoning of the case and give the judgment without any learning, fine-tuning, or examples. On CAIL2018, a Chinese criminal case dataset, we performed zero-shot judgment prediction experiments with GPT-3 models. Our results show that LLMs with LoT achieve better performance than the baseline and chain of thought prompting, the state-of-art prompting method on diverse reasoning tasks. LoT enables the model to concentrate on the key information relevant to the judgment and to correctly understand the legal meaning of acts, as compared to other methods. Our method enables LLMs to predict judgment along with law articles and justification, which significantly enhances the explainability of models.
\end{abstract}



\maketitle

\section{Introduction}

Legal judgment prediction (LJP) seeks to predict the judgment of a legal case based on its fact description. It is important and has long history in the field of AI and Law \cite{sartor2022thirty,ashley2019brief}. Deep learning and pre-trained language models have considerably improved the accuracy of LJP. However, from the standpoint of legal professionals \cite{gorski2021explainable}, this approach lacks explainability. In order to reach a judgment, judges should conduct a three-step process of discovering the law, recognizing the facts, and applying the law to the facts. It is a complex form of deductive reasoning that is commonly known as legal syllogism \cite{posner1990problems}. Current LJP models just provide the final judgment without providing the syllogism reasoning process; hence, they are not explainable. And meanwhile, it requires a relatively large quantity of data annotated by legal experts. These constraints restrict the use of LJP in real-world legal scenarios.

Recently, large language models (LLMs) have shown complex reasoning abilities through chain-of-thought prompting \cite{wei2022chain}. It is highly explainable since it teaches LLMs to generate intermediate reasoning steps with only a few exemplars. Further, \cite{kojima2022large} revealed that LLMs are capable of zero-shot reasoning without any task-specific examples. Compared to supervised learning or fine-tuning, the zero-shot ability of LLMs can significantly lower the cost of data annotation. This has motivated us to implement LLMs and chain-of-thought in the AI and Law fields. 

\begin{figure}[!h]
    \includegraphics[width=0.48\textwidth]{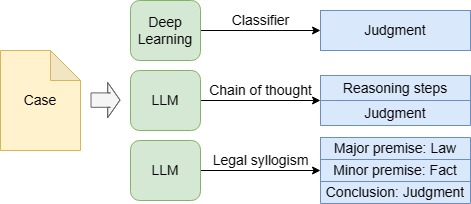}
    \caption{An overview of three different approaches for LJP: (1) Deep learning text classification models for providing the judgment without any explanations. (2) Chain-of-thought prompting for providing the judgment with intermediate reasoning steps as explanations. (3) Legal syllogism prompting for providing the three deductive reasoning steps: law, fact and judgment.}
    \label{fig:intro}
\end{figure}

In this paper, we introduce legal syllogism prompting (LoT), a method for instructing LLMs for zero-shot LJP with legal explainability. As illustrated in Figure 1, we find that LoT can stimulate the deductive reasoning capability of LLMs, which makes it more suitable for the LJP than chain-of-thought. We empirically evaluated LoT on the CAIL2018 dataset \cite{xiao2018cail2018} with the GPT-3 model. The experiment results demonstrate that LoT outperforms chain-of-thought prompting and baseline on the zero-shot LJP task without any learning or fine-tuning. Our work reveals the great potential of LLMs in the fields of AI and law.

\section{RELATED WORK}
\paragraph{\textbf{Legal judgment prediction}} In prior works, LJP is typically viewed as a text classification task based on supervised machine learning, deep learning, or fine-tuning with pretrained models. Katz et al. \cite{katz2017general} introduced random forest to predict the outcome of the Supreme Court of the United States. Medvedeva et al. \cite{medvedeva2018judicial} use support vector machines on case text features to predict the judgment of the European Court of Human rights. Luo et al. \cite{luo2017learning} use the attention neural networks to predict the judgment of criminal cases of the People’s Republic of China. Chalkidis et al. \cite{chalkidis2020legal} explored the BERT model fine-tuned on legal-specific corpora and achieve high performance on downstream case text classification on the European Court of Human rights. Santosh et al. \cite{santosh2022deconfounding} used expert annotations to mitigate the vulnerability of the LJP model. Although these approaches have reached good accuracy, they need a substantial amount of data annotated by legal experts and lack explainability.

\paragraph{\textbf{Large language models and chain of thought}} Large language models (LLMs) are language models trained on a massive amount of text data to predict distribution in a given context. LLMs have strong performance in a wide range of NLP tasks just by prompting methods \cite{liu2023pre}. Trautmann et al. \cite{trautmann2022legal} presented that LLMs can outperform random and majority baseline on judgment prediction of European Court of Human Rights cases.  But they only let the model output a result of Yes or No, without specific prompts to improve reasoning ability. Wei et al. \cite{wei2022chain} proposed a few-shot prompting method, chain of thought, that enables models to generate step-by-step answers to math problems. Kojima et al. \cite{kojima2022large} introduced zero-shot chain-of-thought (Zero-shot CoT) by adding a simple prompt “Let’s think step by step” before each answer. Zero-shot CoT demonstrated the strong reasoning capability of the model without any examples. Wei et al. \cite{wei2022emergent} discussed the multi-steps reasoning ability as an emergent ability, which does not appear in smaller models but appears in larger models. Yu et al. \cite{yu2022legal} showed Zero-shot CoT prompting achieved the best accuracy on COLIEE legal entailment task. It revealed LLMs are capable of legal reasoning. These existing works inspired us how to explore the legal reasoning capabilities of the model with the help of Zero-shot CoT.

\section{legal syllogism prompting}

We propose legal syllogism prompting (LoT), a zero-shot prompting for deductive reasoning in legal NLP tasks. Unlike the original chain of thought, it is zero-shot and does not require additional examples. It is also different from the Zero-shot CoT about the intermediate reasoning steps. The LoT can activate the model's deductive reasoning ability, allowing the model to output the legal reasoning process and conclusion, as shown in Figure 2.

\begin{figure}[t]
    \includegraphics[width=0.48\textwidth]{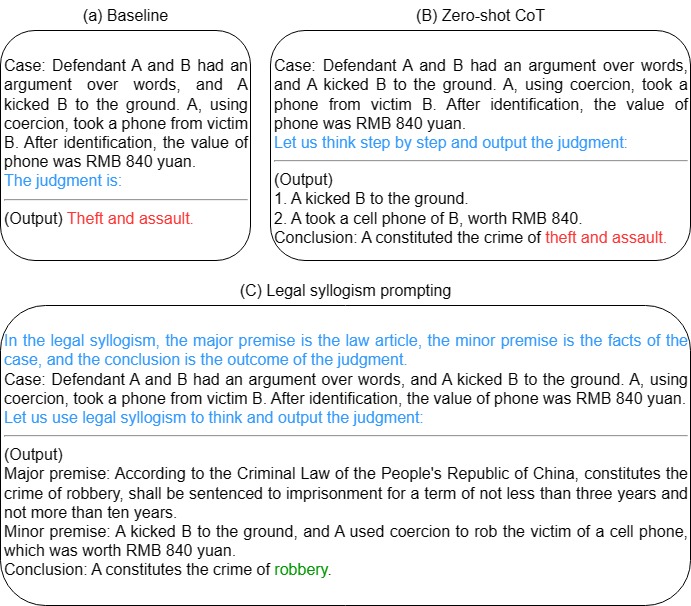}
    \caption{A robbery case as an example to show the difference between three prompting methods to the GPT-3 model: (a) Baseline; (b)Zero-shot CoT; (c)  Our method. Our method output the right judgment using legal syllogism. Despite the steps given, Zero-shot CoT misunderstood the legal meaning of A’s two acts and reached the wrong judgment.}
    \label{fig:syllogism}
\end{figure}

\subsection{Legal syllogism}

Almost every student in law school around the world will meet legal syllogism when they start studying law, regardless of whether it is statutory or case law. Legal syllogism is a specific form of deductive logic in the legal field and can be traced back as far as two thousand years to Aristotle’s classic expression:

\begin{mybox}
\small
Major premise: All men are mortal.

Minor premise: Socrates is a man.

Conclusion: Therefore, Socrates is mortal.

\end{mybox}

In legal judgment, the major premise is the applicable law; the minor premise is the relevant facts; and the conclusion is the judgment. It is a fundamental way of thinking for judges. Early legal scholars usually believed that every case needed to be given a conclusion in the form of a perfect syllogism \cite{huhn2001use}.

This approach has also had a profound impact on AI and law research. Legal AI in the last century built legal rules and legal knowledge into expert systems \cite{governatori2022thirty}. This path fits perfectly into the idea of legal syllogism. However, it still has limitations in terms of efficiency and generalization ability. Modern legal AI commonly uses machine learning to improve accuracy and scalability while losing the ability to syllogistic reasoning. Some researchers have started to use a syllogism-like approach (IRAC, Issue, Rule, Application, and Conclusion) \cite{bench2020explaining} to improve the explainability of machine learning models, but no research has yet combined it with LLMs. These prior works inspired us to combine legal syllogism with LLMs to address these limitations above.

\subsection{Legal syllogism prompting}

The key idea of our approach is simple, as described in Figure 2. The baseline method directly outputs the judgment without any additional instruction. Zero-shot CoT adds "Let us think step by step" as the prompt to model. We let our method gives a brief introduction to legal syllogism before the case fact and then requires the model to use it to give a conclusion. And we reformulate the LJP task from a text classification task to a text generation task. The input fact description of the case is a word sequence: $ \mathbf{X} = \{x_1,x_2,\cdots,x_n\} $ . We add prompts to modify the input $ \mathbf{X} $ into $ \mathbf{X’} $ as the following template:

\begin{mybox}
\small
In the legal syllogism, the major premise is the law article, the minor premise is the facts of the case, and the conclusion is the judgment of case.

Case: $ \mathbf{X} $

Let us use legal syllogism to think and output the judgment:

\end{mybox}

Although we just add some sentences to the original input, we found it can activate the syllogistic reasoning ability of LLMs. We will give a more in-depth review of our approach's benefits in the later section. Our prompt enables LLMs to output relatively fixed format content as below:

\begin{mybox}
\small

Major premise: $\{$law article text$\}$

Minor premise: $\{$fact text$\}$

Conclusion: $\{$legal judgment$\}$

\end{mybox}

Our method can generate both the reasoning process and the conclusion. The reasoning process can be used as an explanation of the conclusion for post hoc verification. After prompting, it only needs to use a simple rule method to process the last paragraph of output to extract the final judgment.

\section{EXPERIMENTS}

\subsection{Dataset}

We evaluated our idea on the Chinese AI and Law challenge dataset (CAIL2018) \cite{xiao2018cail2018}. It is a Chinese criminal case dataset widely used for LJP research.  They are all real cases collected from China Judgments Online, the official  website  that  publishes  cases  and  decisions  from  Chinese  courts. Each case involves both factual description and legal judgment. The judgment further includes three parts: law articles, charges, and prison terms. The charge is the key outcome in criminal cases, so we select the judgment of charge for evaluation.

CAIL2018 involves more than one hundred charges and one hundred thousand data, in which the charges are in an unbalanced distribution. And LLMs have not been trained on specific legal corpus before and have not learned the LJP task. LLMs may not perform well on difficult charges that require more legal knowledge. Therefore, we first ran the baseline method on the CAIL2018 second-stage test set (35k) \cite{zhong2018overview} and observed the performance of the GPT-3 model (text-davinci-003) on different charges. The baseline method requires the model to output the charges directly, as described in Figure 2, without other prompt sentences.
We found the baseline method only achieved the accuracy of about 30$\%$ on the test set. Most errors are due to outputting inaccurate charge names. For example, dangerous driving is a high-frequency charge in criminal cases in China. According to Chinese criminal law, dangerous driving includes drunk driving, overloading, transporting dangerous goods, and other behaviors. Yet LLMs do not know this knowledge and output the judgment as the crime of drunk driving. This leads to low accuracy on a large number of charges.

We aimed to assess the effectiveness of our prompt method, not the performance of LLMs. There is no need to use the entire dataset. Therefore, we selected eight charges that have high frequency in the dataset and have good baseline performance. They are fraud, theft, robbery (Rob), murder (Mud), rape, property damage (Pod), intentional injury (Ini), and negligently serious injury (Nsi). They are also common in the real world. Considering the experiment cost of GPT-3, we randomly sampled 100 examples for each charge from the CAIL2018. The total number of our dataset is 800. It is relatively close to other LLMs prompt experiments \cite{saha2022hard,shi2023large}. We did not use other data to fine-tune LLMs in all experiments. We also did not provide charge names list to models for two considerations: (1) The experiment cost of GPT models is related to the input token numbers. The total charge numbers in CAIL2018 are more than one hundred. (2) If we only provide eight selected charge names, it will provide reference answers that may interfere with the models.

\subsection{Experimental setup}

\paragraph{\textbf{Baselines}} We aimed to analyze the deductive reasoning ability of our legal syllogism prompting to Zero-shot CoT and baseline zero-shot prompting, as showed in Figure 2. We did not compare with original chain of thought, which is a few-shot method and requires manual construction of exemplars. Providing examples will add additional information about the charges, which will help the model identify the cases as same as the samples. It is not fair to cases with other charges. Therefore, we did not select the original chain of thought method. 

\paragraph{\textbf{Metrics}} In previous work \cite{luo2017learning,feng2022legal}, models know the total types of charges when training LJP as a multi-classification task. In our zero-shot experiment, we don’t give LLMs about how many types of charges there are, so the charge names outputted by the model are unstable. We could only calculate the accuracy as other chain-of-thought research \cite{kojima2022large}. The number of cases in each charge is the same. Therefore, We use micro accuracy as our evaluation metrics. The code and sampled data used in this work are released on Github.\footnote{https://github.com/JiangCong7/Legal-Syllogism-Prompting}

\begin{table*}[]
\centering
\caption{Performance of the different prompting methods for LJP on two GPT-3 models. * The outputs of text-davinci-002 are unstable. Therefore, we could not compute the actual accuracy of text-davinci-002, nor we could calculate the accuracy of each charge type.}
\resizebox{0.9\textwidth}{!}{%
\begin{tabular}{cllllllllll}
\hline 
\multicolumn{9}{c}
{CAIL2018 sampled dataset}                                \\ \cline{3-11} 
                  &                   & \multicolumn{1}{l}{Total} & Fraud & Theft & Rob  & Mud  & Rape & Pod   & Ini  & Nsi  \\ \hline
                  & Baseline          & 0.6450                    & 0.67  & 0.92  & 0.41 & 0.58 & 0.92 & 0.61 & 0.94 & 0.07 \\ \cline{2-11} 
Text-davinci-003  & Zero-shot CoT     & 0.5875                    & 0.63  & 0.82  & 0.34 & 0.56 & 0.79 & 0.64 & 0.89 & 0.12 \\ \cline{2-11} 
                  & Legal syllogism   & \textbf{0.6850}                    & 0.70  & 0.94  & 0.45 & 0.64 & 0.95 & 0.75 & 0.91 & 0.15 \\ \hline
\multicolumn{1}{l}{}                    & Baseline                               & 0.1313*                      &                            &                            &                           &                           &                           &                           &                           &                           \\ \cline{2-3}
Text-davinci-002*                       & Zero-shot CoT                          & 0.4925*                      &                            &                            &                           &                           &                           &                           &                           &                           \\ \cline{2-3}
\multicolumn{1}{l}{}                    & Legal syllogism                        & 0.4038*                      &                            &                            &                           &                           &                           &                           &                           &                           \\ \hline
\end{tabular}%
}
\end{table*}

\subsection{Results}
Table 1 summarizes the accuracy of baseline, Zero-shot CoT, and our method (LoT). LoT outperforms the other two methods on CAIL2018 sampled dataset. LoT can also achieve better accuracy in each of the eight legal judgment categories. The accuracy of Zero-shot CoT is lower than the baseline. We think this is because the intermediate reasoning steps generated by Zero-shot CoT do not conform to legal reasoning. We will give a detailed analysis of it in the later section. Our experimental results demonstrate that LoT can successfully enhance the legal reasoning capacity of LLMs. With the prompt of legal syllogism, LLMs can improve the accuracy of LJP without any learning or fine-tuning.

We also conducted comparative experiments on six different GPT models, text-davinci-003/002/001 and text-curie/babbage/ada-001. Except for the text-davinci-003/002 models, other models were not capable of finishing LJP task and did not give any reliable output. Therefore, we do not present the experimental results of the other models in Table 1. For text-davinci-002, the generated outputs were  unstable. For example,  The model outputted hundreds of irrelevant tokens in the LoT method, such as evidence, appeals, trial proceedings, and trial staff. This leaded to inaccurate result of the model. The LLMs have a limit on the length of the output tokens. If it outputs too many irrelevant tokens in the intermediate steps, models may not be able to give the entire output. It is possible that the 
charge name part is not outputted by models, which makes it also difficult to calculate accuracy. We think it is related to the emergent ability \cite{wei2022emergent} of LLMs. Only large enough LLMs have complicated reasoning ability and stable generation capability. And smaller LLMs cannot address the LJP task in the generation approach.

\subsection{Error analysis}
To gain a deeper understanding of the behaviors of our methods, we randomly examined the error results. Our method may output the mistake law article. For example, for a negligent injury case, the model output the major premise as: \textit{Negligent causing serious injury constitutes intentional injury}. This causes the model to give an inaccurate judgment.

Secondly, the numbering of the law article in the major premise generated by our method may be incorrect. For instance, in a robbery case, the model output: \textit{According to Article 266 of the Criminal Law of the People's Republic of China, … constitutes the crime of robbery}. The true number of robbery is \textit{Article 263}.

Although the model finally outputs the correct conclusion, the inaccurate article number may also be misleading and decrease the trustworthiness of the model. Our research focuses on the law-based reasoning capabilities of LLMs and the accuracy of law articles is not our priority issue. We think it could be improved in future works by a post-processing method of introducing external legal knowledge \cite{shuster2022language}. 

\section{DISCUSSION}
In this section, we illustrate the advantages of our approach over Zero-shot CoT and traditional LJP models from different aspects.

\textbf{Selectivity} is the ability to concentrate on relevant parts of making decisions \cite{atkinson2020explanation}. We observed Zero-shot CoT usually considered facts that are not related to legal judgment, in the intermediate step. For a theft case example, Zero-shot CoT outputted as below:

\begin{mybox}
\small
1. A saw B put his phone in his coat bag, and \colorbox{green!30}{took} it when he was not aware of it.

2. A \colorbox{red!30}{surrendered} to the police. 

3. B \colorbox{yellow!30}{issued a letter to express} his understanding of A.

Conclusion: A constitutes the crimes of \colorbox{green!30}{theft} and \colorbox{red!30}{surrender.}

\end{mybox}

It may lead to errors in judgment. For LoT, due to the existence of the major premise, the minor premise will only contain facts related to the law article in the major premise:

\begin{mybox}
\small
Major premise: According to Criminal Law, the crime of \colorbox{green!30}{theft} is …

Minor premise: A saw B put his phone in his coat bag, and \colorbox{green!30}{took} it when he was not aware of it.

Conclusion: A constitutes the crime of \colorbox{green!30}{theft.}

\end{mybox}

\textbf{Sensitivity} refers to the capability to distinguish different charges \cite{an2022charge}. We found that Zero-shot CoT tended to interpret several continuous acts into multiple charges. For a fraud case for example, Zero-shot CoT respectively identified the three fraud acts as three different charges:

\begin{mybox}
\small
1. A \colorbox{green!30}{mistakenly stated} that the village land would be used for mining. A \colorbox{green!30}{fooled} B into buying the land and reselling it to the mining for a great profit.

2. A asked B to \colorbox{red!30}{transfer 500,000 yuan} from his company's accountant to A. 

3. A used them to \colorbox{red!30}{repay his loan} on the same day.

Conclusion: A constitutes the crimes of \colorbox{green!30}{fraud,} \colorbox{red!30}{accepting bribes,} and \colorbox{red!30}{misappropriation of public funds.}
    
\end{mybox}

LoT was sensitive enough and consider these three acts together as the crime of fraud as below:

\begin{mybox}
\small
Major premise: Under Criminal Law of …, the crime of \colorbox{green!30}{fraud} is the \colorbox{green!30}{acquisition of property} with the intent to \colorbox{green!30}{deceive others.}

Minor premise: A \colorbox{green!30}{lied} to B that he could make a profit by acquiring land at a low price and reselling it at a high price, \colorbox{green!30}{deceived} B, and then \colorbox{green!30}{caused B to transfer 500,000 yuan} to A.

Conclusion: A constitutes the crimes of \colorbox{green!30}{fraud.}

\end{mybox}

\textbf{Explainability and Generality} Both Zero-shot CoT and LoT can enhance explainability. However, Zero-shot CoT explains the conclusion with a step-by-step interpretation of facts. The law and rules are rarely included. LoT analyzes both the law and the facts and is more consistent with legal logic. On the other hand, previous works typically considered the prediction of law article and judgment as two different subtasks. LoT can provide both law articles and judgments, which gives the model a stronger general ability.

\section{CONCLUSIONs and future work}

In this paper, we propose LoT, a new prompting method to teach LLMs to predict case judgment. We apply an old-school theory, legal syllogism, in cutting-edge AI models. This combination presents the potential for integrating AI and legal knowledge. However, we must also point out the limitations of our methods.
In legal syllogism, in spite of the deductive reasoning from premise to conclusion, what is more critical is the process of leading to the major and minor premise. In this process, judges will usually interpret the law and reconstruct the facts. This is often referred to as practical reasoning \cite{posner1990problems}. After confirming that LLMs could learn syllogism, whether it has the ability of practical reasoning is an important research direction. We anticipate that our work will encourage more researchers to investigate the reasoning abilities of LLMs on legal NLP tasks and combine legal theory with AI techniques.

\section{ETHICAL CONSIDERATION}
Our work could assist the community of AI and Law in better understanding the reasoning ability of LLMs. We demonstrate that the judgment generated by LLMs still contains many errors, and we recommend using these models and prompting methods with greater care and consideration. We think the goal of LJP is not to replace judges, but rather to support them and other legal communities, so enhancing the effectiveness of the legal system and reducing the cost of legal resources.

\begin{acks}
We appreciate the valuable feedback by the anonymous reviewers. This work was supported by Peking University - Wuhan East Lake Intelligent Social Governance Opening Program titled "Intelligent public legal consulting service system of East Lake High-Tech Development Zone" (ZX2222M).
\end{acks}

\bibliographystyle{ACM-Reference-Format}
\bibliography{sample-base}

\end{document}